\documentclass[review]{elsarticle}

\usepackage{lineno,hyperref}
\usepackage{algcompatible}
\usepackage{algorithm}
\usepackage[shortlabels]{enumitem}
\usepackage{epstopdf}
\usepackage{dsfont}
\usepackage{graphicx}
\usepackage{subfigure}
\usepackage[section]{placeins}
\usepackage{multirow}
\usepackage{hyperref}
\usepackage{amsthm}
\usepackage{amsfonts}
\usepackage{xcolor}
\usepackage{natbib} 
\usepackage{cool}
\usepackage{centernot}
\usepackage[T1]{fontenc}
\usepackage[left=2.5cm,right=2.5cm, bottom = 2.2cm, top = 2.2cm]{geometry}

\newtheorem{Def}{Definition}
\newtheorem{theorem}{Theorem}
\newtheorem{remark}{Remark}
\newtheorem{corollary}{Corollary}

\newtheorem*{Ass}{Assumptions}
\newtheorem*{Cond}{Conditions}

\newcommand{\R}{{\mathbb{R}}}
\newcommand{\N}{{\mathbb{N}}}

\modulolinenumbers[5]

\journal{Journal of \LaTeX\ Templates}

\begin{document}

\begin{frontmatter}

\title{Interpretable Machines: Constructing Valid Prediction Intervals with Random Forests}

\author{Burim Ramosaj$^*$}
\address{Department of Statistics \\
	 Institute of Mathematical Statistics and Applications in Industry\\
	 TU Dortmund University \\
	44227 Dortmund, Germany}


\cortext[mycorrespondingauthor]{\textit{Corresponding Author:} Burim Ramosaj \\
	\textit{Email address:} \texttt{burim.ramosaj@tu-dortmund.de} }


\begin{abstract}
	An important issue when using Machine Learning algorithms in recent research is the lack of interpretability. Although these algorithms provide accurate point predictions for various learning problems, uncertainty estimates connected with point predictions are rather sparse. A contribution to this gap for the Random Forest Regression Learner is presented here. Based on its Out-of-Bag procedure, several parametric and non-parametric prediction intervals are provided for Random Forest point predictions and theoretical guarantees for its correct coverage probability is delivered. In a second part, a thorough investigation through Monte-Carlo simulation is conducted evaluating the performance of the proposed methods from three aspects: (i) Analyzing the correct coverage rate of the proposed prediction intervals, (ii) Inspecting interval width and (iii) Verifying the competitiveness of the proposed intervals with existing methods. The simulation yields that the proposed prediction intervals are robust towards non-normal residual distributions and are competitive by providing correct coverage rates and comparably narrow interval lengths, even for comparably small samples. 
\end{abstract}


\end{frontmatter}


\section{Introduction}

Several prediction tasks using bagging and boosting procedures among Machine Learning (ML) algorithms have revealed favourable point predictions. Examples can be found in \cite{buhlmann2003bagging}, \cite{prasad2006newer}, \cite{jiang2007mipred}, \cite{pham2018bagging} or \cite{wang2018random}, where point predictions have been constructed accurately when measured through $L_2$-error loss, the area-under-the-ROC-curve (AUC), the mis-classification error or other measures for evaluating prediction accuracy. The restriction to bagging and boosting as a class of ML algorithms has several advantages: the methods make use of easy to construct weak learners such as decision trees and aggregate them through a weighted voting scheme or by introducing bootstrap methods. They usually result into methods, that have few hyperparameters to tune and can be trained in reasonable time complexity, while significantly increasing prediction power compared to their single learners. Especially the Random Forest method, which is an ensemble of randomized trees, have shown favourable results from three aspects: accurate point predictions, less tuning efforts and the ability to select \textit{important features}, see e.g. \cite{breiman2001random}, \cite{diaz2006gene}, \cite{jiang2007mipred}, \cite{genuer2010variable} or \cite{ramosaj2019asymptotic} . These trends have made recent research in statistical ML to focus on the derivation of statistical properties for such models. In \cite{wager2014asymptotic}, for example, the authors proposed a modified Random Forest model in order to establish consistency and derive central limit type theorems for these modified Random Forest models. In  \cite{scornet2015consistency}, for example, the authors focused on the original version of Breiman's Random Forest \cite{breiman2001random} and could show under conditions such as additive regression functions that the Random Forest regression learner is $L_2$-consistent. In several other authors works such as in \cite{mentch2016quantifying} or \cite{zhou2019v}, for example, a connection between a Random Forest learner and incomplete $U$-statistics have been established. This, for the reason to make use of established central limit type theorems in order to ease the way to statistical testing procedures. The latter is closely connected to the quantification of uncertainty in point predictions, see e.g. \cite{wager2014confidence}. The theoretical work conducted so far can have severe implications to the general use of ML-based methods. While applicants of such methods have criticized ML-based methods for being \textit{not interpretable} in terms of lacking the involvement of uncertainty estimates in prediction tasks, the recent research trend indicates that \textit{interpretable ML} is an active issue in statistical ML, see e.g. \cite{adadi2018peeking}, \cite{dunson2018statistics}, \cite{guidotti2018survey} or  \cite{molnar2020InterpretableMachine}. \\

It is aimed to contribute to the general issue of interpretable ML from a statistical perspective through the construction of valid prediction intervals, that are mainly based on the Random Forest method. The construction of such intervals enables an answer to the general question researchers and practitioners have been stating \cite{dunson2018statistics}: \textit{How trustworthy is the ML-based algorithm in its point-prediction for future observations?} The access to a certain range for point-predictions within a given certainty delivers a potential awnser to this question. The construction of such intervals, however, requires statistical properties such as consistency of the underlying ML algorithm. As mentioned previously, consistency results have already been established such that the gap towards the construction of point prediction intervals can be settled up based on these findings. Regarding the construction of prediction intervals using the Random Forest, recent findings have been made in \cite{coulston2016approximating}, \cite{calvino2020random} or \cite{roy2020prediction}. Therein, the authors mainly derive point prediction intervals based on the Random Forest from a simulation based perspective. In \cite{coulston2016approximating}, for example, the authors recommended the construction of uncertainty estimates in geographical maps, for e.g. forest fire behaviour. In \cite{calvino2020random}, a special focus has been put on Quantile Regression Forest, the linear quantile regression and modifications of the Random Forest. Therein, the authors conducted numerical experiments and recommended not to use Quantile Regression Forests for the construciton of point-prediction intervals. Regarding the latter, a theoretical work has been given in \cite{meinshausen2006quantile}, where the authors delivered theoretical guarantees for the validity of quantile based prediction intervals using the Random Forest. However, assumptions regarding the tree construction process have been set, that are rarely met when using Breiman's Random Forest method \cite{breiman2001random} as implemented in the \textsf{R}-package \texttt{randomForest}. Therefore, obtaining correct coverage rates under the Quantile Regression Forest scheme as given in \cite{meinshausen2006quantile} remains unclear and will be investigated within an extensive simulation study in this work. In addition to the stated examples, the authors in \cite{roy2020prediction} proposed several prediction intervals distinguishing between parametric- and non-parametric designs. Theoretical guarantees for its correct coverage are still lacking there. This is in contrast to the work of \cite{zhang2019random}, where the authors constructed both, a theoretical framework for the validity of Random Forest based prediction intervals. In its core, they propose to use empirical quantiles of the Random Forest method that has been computed on the Out-of-Bag set. Under the assumptions of $L_2$-consistent Out-of-Bag Random Forests, they deliver theoretical guarantees for several types of coverage rates for prediction intervals. \\

This work is focused on the derivation of theoretical guarantees for Random Forest based prediction intervals that have been i) constructed parametrically and ii) non-parametrically. Compared to the work of \cite{zhang2019random}, assumptions are relaxed while still maintaining correct coverage rates. Furthermore, an additional Random Forest Out-of-Bag prediction interval is proposed, that is of parametric nature and give theoretical guarantees for their correct (asymptotic) coverage. In its core, it is based on consistently estimating residual variance in regression learning problems. The latter can be considered as a non-trivial underpinning and has been partly investigated in e.g. \cite{ramosaj2019consistent}. Additional residual variance estimators are proposed that correct for the finite choices of decision trees in the ensemble. In addition to these findings, an extensive simulation study is conducted indicating the good performance of the proposed method for obtaining parametric prediction intervals based on the Random Forest Out-of-Bag prediction. Therein, both, coverage rates and interval lengths for several types of prediction intervals were analyzed, as classified in \cite{zhang2019random}.\\

This work is structured as follows: In Section \ref{Sec:ModelFrame}, the general model framework is introduced, which is restricted to the univariate-response regression. A general, theoretical framework is derived, under which (asymptotically) valid prediction intervals can be obtained and a formal definition for different types of prediction intervals is set, similar to \cite{zhang2019random}. In Section \ref{Sec:RandomForest}, the Random Forest regression method and the idea of using Out-of-Bag observations in the forest is shortly introduced. Furthermore different types of residual variance estimators are established and their consistency is proven. This way, potential sources having impact on the speed of converges are identified and corrected for. Based on these findings, several parametric prediction intervals are proposed. In Section \ref{Sec:AlternativeML}, alternative methods such as the XGBoost or the Stochastic Gradient Tree Boosting method together with the Quantile Regression Forest method are considered for potential benchmarking with Random Forest based prediction intervals. The idea is to establish an analogy to the Random Forest based prediction intervals to boosting methods. The simulation design and framework are then presented in Section \ref{Sec:Simulation}, while Section \ref{Sec:SimResults} briefly summarize the main findings in this simulation. In the last section, the main results are summarized and a short overview on future research in  this field is given. Note that this paper consists of supplementary material covering all proofs for the theoretical guarantees and additional simulation results.

\section{Model Framework and Valid Prediction Intervals} \label{Sec:ModelFrame}

Throughout the paper, assume that one has access to a set $\mathcal{D}_n = \{ [\boldsymbol{X}_i^\top, Y_i]^\top \in \R^{n \times (p+1) } : i = 1, \dots, n  \}$ consisting of iid random vectors with metric outcome $Y \stackrel{d}{=} Y_1 \in \R$. Assuming furthermore that the relation between the response $Y$ and the covariate $\boldsymbol{X} \stackrel{d}{=} \boldsymbol{X}_1$ can be described through a measurable function $m: \R^p \longrightarrow \R$ with 
\begin{align}\label{RegModel}
    Y_i &= m(\boldsymbol{X}_i) + \epsilon_i
\end{align}
for all $i = 1, \dots, n$, where $\{ \epsilon_i \}_{i = 1}^n$ is a sequence of iid random variables with $\mathbb{E}[\epsilon_1] = 0$ and $Var(\epsilon_1) = \sigma^2 \in (0, \infty)$ independent of $\{ \boldsymbol{X}_i \}_{i = 1}^n$. Prediction intervals usually cover the aspect of deriving a measurable region in form of an interval, for which the target at new and unseen feature input $\boldsymbol{X}_0$, i.e. $Y(\boldsymbol{X}_0)$, lies in with a pre-specified certainty, say $1-\alpha \in (0,1)$. In order to formally define the random interval-type set, it is required to distinguish, similarly to \cite{zhang2019random}, between different types of prediction intervals. Note that the obtained interval is potentially random and it is denoted by $\mathcal{C}_{n,1 - \alpha} = \mathcal{C}(\mathcal{D}_n, 1-\alpha)$ to emphasize its dependence towards the data set $\mathcal{D}_n$ and the error rate $\alpha$. 

\begin{Def}\label{PI_Definition}
Consider the regression model in $(\ref{RegModel})$ and fix $\alpha \in (0,1)$. Assume furthermore that $\boldsymbol{X}_0 \stackrel{d}{=} \boldsymbol{X}_1 $ is unseen. The interval-type set $\mathcal{C}_{n, 1 - \alpha}$ is called a prediction interval of 
\begin{enumerate}
    \item \textbf{type-I}, if $\mathbb{P}[ Y(\boldsymbol{X}_0) \in \mathcal{C}_{n, 1-\alpha} ] \ge 1 - \alpha$ holds. 
    \item \textbf{type-II}, if $\mathbb{P}[ Y(\boldsymbol{X}_0)  \in \mathcal{C}_{n, 1 - \alpha} | \mathcal{D}_n] \ge  1- \alpha$ holds almost surely. 
    \item \textbf{type-III}, if $\mathbb{P}[ Y(\boldsymbol{X}_0) \in \mathcal{C}_{n, 1 - \alpha}  | \boldsymbol{X}_0 = \boldsymbol{x}_0 ] \ge 1 - \alpha$ holds for a fixed $\boldsymbol{x}_0 \in \R^p.$
    \item \textbf{type-IV}, if $\mathbb{P}[ Y(\boldsymbol{X}_0) \in \mathcal{C}_{n,  1- \alpha} | \mathcal{D}_n, \boldsymbol{X}_0 = \boldsymbol{x}_0 ] \ge 1 - \alpha$ holds almost surely.  
\end{enumerate}
The prediction interval is said to be asymptotically type-I, type-II, type-III or type-IV valid, if the respective relation above holds for an increasing sample size $n \rightarrow \infty$, while the almost sure statements are substituted by convergence in probability. 
\end{Def}

In practice, the type-IV prediction interval represent the most realistic quantification of uncertainty regarding future prediction points. This, because the obtained interval $\mathcal{C}_{n, 1 - \alpha}$ is usually computed on an observed sample $\mathcal{D}_n$, while the future point to be predicted is also treated as fixed $\boldsymbol{X}_0 = \boldsymbol{x}_0$. Note that a prediction interval of type-IV is also a prediction interval of type-III but not vice versa. A similar relation also holds between a type-I and type-II prediction interval. The latter implies the validity of the type-I condition, but not vice-versa. Furthermore, the type-III prediction interval also implies the type-I condition resulting into a hierarchical system of implications. Therefore, the type-I prediction interval can be considered as the weakest form of a prediction interval. This, because an averaging also happens on the future prediction inputs $\boldsymbol{X}_0$ and makes a verbal interpretation more difficult. Instead, it can be considered as an \textit{averaged prediction interval}, in which, on average, future observations will fall in with a probability of at least $1 - \alpha$.  \\
In order to specify the structure of the interval more into detail, one distinguish between \textit{parametric and non-parametric intervals}. The upcoming sections state general conditions, under which prediction coverage rates as given in Definition \ref{PI_Definition} are valid. Regarding non-parametric prediction intervals, conditions for the validity of Definition \ref{PI_Definition} for the Random Forest method have been given in \cite{zhang2019random}.

\subsection{Parametric Prediction Intervals}

Let us start by setting conditions, for which type-I prediction intervals are valid. Therefore, one denotes with $\widehat{m}_n(\boldsymbol{X}_0)$ any type of estimator, that has been trained on $\mathcal{D}_n$ and predicts the outcome at $\boldsymbol{X}_0$. One can guarantee the asymptotic type-I coverage, if, for example, the following conditions hold:

\begin{Cond}[Parametric Intervals] 
\text{ \\ }
\begin{enumerate}[label=({P\arabic*})]
    \item The learning method is consistent in $\mathbb{P}$-sense, i.e. $\widehat{m}_n(\boldsymbol{X}_0) \label{ParamCond1} \stackrel{\mathbb{P}}{\longrightarrow} m(\boldsymbol{X}_0)$, as $n \rightarrow \infty$. \label{ParamCond2}
    \item The errors $\{ \epsilon_i \}_i$ are Gaussian with mean $0$ and variance $\sigma^2$. \label{ParamCond2}
    \item Any type of residual variance estimator $\widehat{\sigma}_n^2$ is $\mathbb{P}$-consistent, i.e. $\widehat{\sigma}_n^2 \stackrel{\mathbb{P}}{\longrightarrow} \sigma^2$, as $n \rightarrow \infty$. \label{ParamCond3}
\end{enumerate}
\end{Cond}

The assumptions of Gaussian error rates can be relaxed by assuming a certain other limiting, (parametric) distribution for the standardized residuals $\{ \epsilon_i / \sigma \}_{i = 1}^n$ that is continuous. In order to have type-III or type-IV prediction coverage, it is required to substitute the $\mathbb{P}$-consistency of the first condition to a pointwise consistency for fixed $\boldsymbol{x}_0 = \boldsymbol{X}_0$. Note that the above conditions do not deal with heteroscedastic residual variance in the form of $\sigma^2 = \sigma^2(\boldsymbol{x}_0)$. The latter will not be covered in this setting. Therefore, any type of learning algorithm that fulfills conditions $\ref{ParamCond1}$ - $\ref{ParamCond3}$ will lead to a type-I prediction interval that is asymptotically valid. The corresponding type-I prediction interval has then the following form
\begin{align}\label{Param_Interval}
    \mathcal{C}_{n, 1- \alpha} = [ \widehat{m}_n(\boldsymbol{X}_0) + q_{\alpha/2} \cdot \widehat{\sigma}_n, \quad \widehat{m}_n(\boldsymbol{X}_0) + q_{1-\alpha/2} \cdot \widehat{\sigma}_n ],
\end{align}

where $q_{\alpha/2}$ denotes the corresponding $\alpha/2$-quantile of the (standardized) residual distribution. A formal proof for the validity of this parametric prediction interval is given in \cite{ramosaj2020Dissertation} on page $51$, for example. 

\subsection{Non-Parametric Prediction Intervals}

Non-parametric prediction intervals do not make any assumptions on the residual distribution, except the conditions arising from the regression model in $(\ref{RegModel})$. Therefore, assumption $\ref{ParamCond2}$ is usually dropped and substituted by any residual distribution that is continuous. This scenario has been considered in \cite{zhang2019random}, where the authors revealed under the mentioned conditions the asymptotic type-I validity of the prediction interval given by 
\begin{align}\label{NonParam_Interval}
    \mathcal{C}_{n, 1-\alpha}^{NP} &= [ \widehat{m}_n(\boldsymbol{X}_0) + \widehat{D}_{n, \alpha/2}, \quad \widehat{m}_n(\boldsymbol{X}_0) + \widehat{D}_{n, 1 - \alpha/2} ].
\end{align}
The interval length is mainly driven by the empirical quantiles  $\widehat{D}_{n, \alpha}$, which is computed based on the empirical residuals $ \widehat{\epsilon}_{i, n} := Y_i - \widehat{m}_n(\boldsymbol{X}_i)$ using the $\lceil \alpha \cdot n \rceil$-smallest observation among the order statistic $\widehat{\epsilon}_{(1), n},\widehat{\epsilon}_{(2), n} \dots, \widehat{\epsilon}_{(n), n }$. Slight modifications to the computed residuals do exists such as the split-conformal prediction interval, where the absolute residuals are considered among a prior training set seperation. For details on this, it is recommended to check \cite{lei2018distribution} and \cite{zhang2019random}. Similarly to the previous arguments, type-III and type-IV coverage can be obtained also for $\mathcal{C}_{n, 1 - \alpha}^{NP}$, if assumption $\ref{ParamCond1}$ is substituted to the pointwise convergence for fixed $\boldsymbol{x}_0 = \boldsymbol{X}_0$. The benefits of relaxing the Gaussian distribution by any, potentially non-parametric distribution with continuous distribution function expects to have a certain drawback: the interval length might be larger compared to its parametric counterpart. However, this effect needs to be analyzed more thoroughly and will be considered in the simulation Sections \ref{Sec:Simulation} and \ref{Sec:SimResults}. Beside the prediction interval $\mathcal{C}_{n, 1-\alpha }^{NP}$, a Random Forest based type-III asymptotic prediction interval was proposed in \cite{meinshausen2006quantile} mainly focusing on the quantiles estimating directly on obtained response values $\{ \widehat{Y}_i(\boldsymbol{x}_0) \}_{i}$. A more detailed description on this will be delivered in Section \ref{Sec:AlternativeML} under the quantile regression scheme.

\section{Random Forest Regression}\label{Sec:RandomForest}

Random Forest Regression according to \cite{breiman2001random} is an ensemble of randomized decision trees combining elements of bagging and feature sub-spacing. The final aggregation and therefore prediction is conducted by averaging the target variable falling in the same node of the tree as the input $\boldsymbol{X}_0$. For a detailed algorithmic description of the Random Forest Regression, the works \cite{breiman2001random} or \cite{scornet2015consistency} are recommended. However, the important hyper-parameters for the Random Forest learner are shortly listed: 
\begin{itemize}
\item The number of decision trees $M \in \N$. 
\item The cardinality of the feature-subspacing procedure at every node, i.e. $m_{try} \in \N  \cap \{ 1, \dots, p \} $. 
\item The number of samples points $a_n \in \{1, \dots, n\}$ in the bagging step.
\item The number of terminal nodes $t_n \in \N$ each decision tree can have. 
\end{itemize}
The method enables the construction of internal accuracy measures through the usage of the \textbf{Out-of-Bag} scheme. Hence, a prediction at an input $\boldsymbol{X}$ potentially being part of the training set $\mathcal{D}_n$ is computed along all those decision trees in the ensemble, where $\boldsymbol{X}$ has not been part of the tree construction process. This is possible, since Random Forest is a bagging learner resulting into decision trees, that do not include all observational points during the tree construction process. The Out-of-Bag scheme is especially used to obtain internal accuracy measures regarding prediction performance of the Random Forest. Although the Out-of-Bag scheme does not primarily impact the point prediction of the Random Forest method, it will play a key role in the estimation of the residual variance in parametric prediction intervals. This, because the unseen input $\boldsymbol{X}_0$ is not part of the training set making the usage of all trees in the forest for prediction possible. Therefore, one defines a Random-Forest Out-of-Bag prediction at input $\boldsymbol{X}_i$ for a fixed $i \in \{1, \dots, n \}$ using $M$ decision trees as $m_{n, M}^{OOB}(\boldsymbol{X}_i)$ and the corresponding prediction at any other prediction point $\boldsymbol{X}_0$ not included in $\mathcal{D}_n$ as $m_{n, M}(\boldsymbol{X}_0)$. Note that $ m_{n, M}(\boldsymbol{X}_0) = m_{n, M}^{OOB}(\boldsymbol{X}_0)$. One furthermore distinguishes among these point estimators by introducing a theoretically infinite number of decision trees, i.e. $m_{n, \infty}(\boldsymbol{X}_0) = \lim\limits_{M \rightarrow \infty} m_{n, M}(\boldsymbol{X}_0)$ is the Random Forest prediction at $\boldsymbol{X}_0$ using an infinite number of decision trees. Regarding its Out-of-Bag counterpart with an infinite number of decision trees denoted as $m_{n, \infty}^{OOB}(\boldsymbol{X}_i)$, it is proven in the Supplement (Proposition 1) that it exits and attains a conditional expectation form. Any bias introduced through the usage of a finite number of decision trees is called as the \textit{finite-M-bias} theoretically given by $m_{n, M}(\boldsymbol{X}_0)  - m_{n, \infty}(\boldsymbol{X}_0)$. That the latter is a serious source of prediction accuracy loss has been identified in \cite{wager2014confidence} and \cite{ramosaj2020Dissertation}. This, because theoretical results such as Central Limit Theorems or consistency results are based on the infinite Random Forest. The latter results will be used in the construction of valid residual variance estimators. \\

The construction of Random Forest based  type-I prediction intervals of the form as in $(\ref{Param_Interval})$ requires for its correct coverage the fulfillment of the conditions $\ref{ParamCond1} - \ref{ParamCond3}$. Regarding condition $\ref{ParamCond1}$, the authors in \cite{scornet2015consistency} established a formal proof for the infinite Random Forest $m_{n, \infty}$ under regression functions being continuous and additive, while side conditions on the tuning parameters such as $t_n \rightarrow\infty$, $a_n \rightarrow \infty$ and $t_n(\log(a_n))^9/a_n \rightarrow 0$ for $n \rightarrow \infty$ have been set. In addition to that, Gaussian error terms are assumed leading to the same condition as in $\ref{ParamCond2}$. One is therefore left with condition $\ref{ParamCond3}$, which mainly focuses on consistent residual variance estimators.

\subsection{Consistent Residual Variance Estimators}

The issue of consistently estimating residual variance using Random Forest Out-of-Bag predictions is relatively new. In \cite{ramosaj2019consistent}, for example, the issue has been tackled for the infinite Out-of-Bag Random Forest focusing on sample variance estimates on the Out-of-Bag residuals $\widehat{\epsilon}_{i,n, \infty}^{OOB} = Y_i - m_{n, \infty}^{OOB}(\boldsymbol{X}_i)$ leading to the natural estimate of the form 
\begin{align}\label{InfiniteResVar}
    \widehat{\sigma}_{n, \infty}^2 = \frac{1}{n-1} \sum\limits_{i = 1}^n ( \widehat{\epsilon}_{i, n, \infty}^{OOB}  - \bar{\epsilon}_{\cdot, n, \infty}^{OOB} )^2. 
\end{align}
However, a central practical drawback of this type of estimator is the infinite number of decision trees used in the Random Forest model. In practice, only a finite number of decision trees can be computed leading to a potential bias for a finite choice of $M$. Therefore, a theoretical framework is established, which allows the introduction of a finite-M-bias corrected residual variance estimator. In doing so, the following conditions are set: 
\begin{Ass}
\text{\\}
\begin{enumerate}[label=({A\arabic*})]
    \item The support of the feature input is the p-dimensional hyper-rectangular unit-cell, i.e. $supp(\mathbf{X}) = [0,1]^p$.\label{Ass1}
    \item The regression function is bounded, i.e. $|| m ||_{\infty} = \sup\limits_{\boldsymbol{x} \in  [0,1]^p } |m(\boldsymbol{x})| =: K < \infty$. \label{Ass2}
    \item The residuals are centered Gaussian as in condition $\ref{ParamCond2}$. \label{Ass3}
    \item The Random-Forest estimator $m_{n, \infty}(\boldsymbol{X}_0)$ is $L_2$ consistent, i.e. $\lim\limits_{n \rightarrow \infty}\mathbb{E}[(m_{n, \infty}(\boldsymbol{X}_0) - m(\boldsymbol{X}_0))^2] = 0$.  \label{Ass4}
\end{enumerate}
\end{Ass}

Note that assumption $\ref{Ass1}$ is not severe when using the Random Forest method. The reason is that the latter is invariant to monotone transformation (see e.g. \cite{ramosaj2020Dissertation}, page $26$) including this way a rich class of distributions for the features $\{ \boldsymbol{X}_i \}_{i=1}^n$. Assumption $\ref{Ass3}$ is in line with the assumptions given in \cite{scornet2015consistency} in order to obtain consistency as required in $\ref{Ass4}$. Note that the latter assumption does not imply the consistency of the finite-$M$ Out-of-Bag Random Forest estimator $m_{n, M}^{OOB}$. The latter, however, was treated in \cite{ramosaj2019consistent} and guarantees the consistency of the Out-of-Bag Random Forest under $\ref{Ass4}$ and the stated model framework in $(\ref{RegModel})$. Regarding the consistency of the residual variance estimator in $(\ref{InfiniteResVar})$, only assumptions $\ref{Ass1}$ and $\ref{Ass4}$ within regression model $(\ref{RegModel})$ together with $\mathbb{E}[|m(\boldsymbol{X}_1)|^2]< \infty$ is required, in order to have $\widehat{\sigma}_{n, \infty}^2 \stackrel{L_1}{\longrightarrow} \sigma^2$ as $n \rightarrow \infty$. This has been shown in Theorem $1$ in \cite{ramosaj2019consistent} and is not part of this work. Instead, the focus will be on the derivation of residual variance estimators, that correct for potential bias due to the finite choice of $M$. Denoting with $(n,M) \stackrel{seq}{\rightarrow} a$ the sequential limit of the form $\lim\limits_{n \rightarrow a} \lim\limits_{M \rightarrow a}$, the first theoretical result in the following Theorem is stated enabling the construction of parametric prediction intervals. 

\begin{theorem} \label{MainTheorem}
Assume the regression model $(\ref{RegModel})$ together with the conditions $\ref{Ass1}$, $\ref{Ass2}$ and $\ref{Ass4}$. Then the estimator 
$$ \widehat{\sigma}_{n, M}^2 = \frac{1}{n} \sum\limits_{i = 1}^n  (\widehat{\epsilon}_{i, n, M}^{OOB})^2 $$
with $ \epsilon_{i, n, M} := Y_i - m_{n, M}^{OOB}(\boldsymbol{X}_i)$ is consistent, i.e. 
$$ \widehat{\sigma}_{n, M}^2 \stackrel{L_1}{ \longrightarrow} \sigma^2, $$
as $(n, M) \stackrel{seq}{\rightarrow} \infty$. \\
If in addition $\ref{Ass3}$ holds, then the finite-M bias of the residual variance estimator $\widehat{\sigma}_{n, M}^2$ can be bounded by 
$$ \mathbb{E}[ \widehat{\sigma}_{n, M}^2 - \widehat{\sigma}_{n, \infty}^2 ] \le \frac{8}{M}( || m ||_{\infty} + \sigma^2( 1 + 4\log(n) ) ).  $$
Under these conditions, the finite-$M$ corrected estimator 
\begin{align}
    \widehat{\sigma}_{n, Mcorrect}^2 = \left| \widehat{\sigma}_{n, M}^2  - \frac{8}{M}\left(  ( \max\limits_{1 \le i \le n} |m_{n, M}^{OOB}(\boldsymbol{X}_i)| )^2  + \widehat{\sigma}_{n, M}^2 ( 1 + 4\log(n) )  \right) \right|.
\end{align}
is then $L_1$-consistent for $\sigma^2$, as $(n,M) \stackrel{seq}{\rightarrow} \infty$.
\end{theorem}
The proposed finite-$M$ corrected residual variance estimator $\widehat{\sigma}_{n,M}^2$ for $\sigma^2$ is more conservative than $\widehat{\sigma}_{n,M}^2$ due to the substraction of an upper bound. In practice, it can result in the underestimation of the true residual variance, while $\widehat{\sigma}_{n, M}^2$ usually overestimates the true residual variance. The over- and underestimation of the corresponding residual variance estimators can be used to construct an additional estimator that smooths out this effect. 

\begin{corollary}
Let $\lambda_1 \in (0,1)$ and define $\lambda_2 = 1 - \lambda_1$. Assume that regression model $(\ref{RegModel})$ is valid together with the conditions $\ref{Ass1}$ - $\ref{Ass4}$. Then the estimator
\begin{align*}
    \widehat{\sigma}_{n, M; W}^{2} &= \lambda_1 \widehat{\sigma}_{n,MCorrect}^2 + \lambda_2 \widehat{\sigma}_{n, M}^2 
\end{align*}
is $L_1$-consistent, as $(n, M) \stackrel{seq}{\rightarrow } \infty$. 
\end{corollary}

One refers to the above residual variance estimator as the \textit{weighted, finite-$M$-corrected residual variance estimator}. Combining these findings leads to the proposal of four parametric prediction intervals of type-I using the Random Forest Out-of-Bag principle.

\begin{corollary}
Assume condition $\ref{Ass1}$, $\ref{Ass3}$ and $\ref{Ass4}$. Then, the interval
\begin{align*}
    \mathcal{C}_{n, 1 - \alpha, \infty} &:= [ m_{n, \infty}(\boldsymbol{X}_0) - z_{1-\alpha/2} \cdot \widehat{\sigma}_{n, \infty}, \quad m_{n, \infty}(\boldsymbol{X}_0) + z_{1-\alpha/2} \cdot \widehat{\sigma}_{n, \infty} ]
\end{align*}
is an asymptotically valid  type-I interval, if $\mathbb{E}[|m(\boldsymbol{X}_1)|^2] < \infty$ holds. If instead $\ref{Ass1} - \ref{Ass4}$ holds, then the intervals 
\begin{align*}
    \mathcal{C}_{n, 1 - \alpha, M} &:= [ m_{n, M}(\boldsymbol{X}_0) - z_{1 - \alpha/2 } \cdot  \widehat{\sigma}_{n, M}^2, \quad m_{n, M}(\boldsymbol{X}_0) + z_{1 - \alpha/2} \cdot \widehat{\sigma}_{n, M}^2, ],  \\
    \mathcal{C}_{n, 1-\alpha, M}^{Cor} &:= [ m_{n, M}(\boldsymbol{X}_0) - z_{1-\alpha/2} \cdot \widehat{\sigma}_{n, MCorrect}, \quad m_{n, M}(\boldsymbol{X}_0) + z_{1-\alpha/2} \cdot \widehat{\sigma}_{n, MCorrect} ] \text{ and } \\
    \mathcal{C}_{n, 1-\alpha, M}^{W} &:= [ m_{n, M}(\boldsymbol{X}_0) - z_{1-\alpha/2} \cdot \widehat{\sigma}_{n, M; W}^{2}, \quad m_{n, M}(\boldsymbol{X}_0) + z_{1-\alpha/2} \cdot \widehat{\sigma}_{n, M; W}^{2} ]
\end{align*}
are asymptotically type-I prediction intervals for $(n, M) \stackrel{seq}{\rightarrow} \infty$ correcting for finite-M bias. 
\end{corollary}

\begin{remark}
All three parametric prediction intervals $\mathcal{C}_{n, 1- \alpha, \infty}$, $\mathcal{C}_{n, 1-\alpha, M}$ and $\mathcal{C}_{n, 1- \alpha, M}^{W}$ can be extended to type-III and type-IV prediction intervals, if assumption $\ref{Ass4}$ is replaced by a stronger argument such as point-wise consistency, i.e. $\lim\limits_{n \rightarrow \infty} \mathbb{E}[ (m_{n, \infty}(\boldsymbol{X}_0) - m(\boldsymbol{X}_0) )^2  |  \boldsymbol{X}_0 = \boldsymbol{x}_0 ] = 0$, depending on the conditioning as given in Definition \ref{PI_Definition}.
\end{remark}

\section{ Other Learning Methods } \label{Sec:AlternativeML}

It is interesting to know whether the general assumptions in $\ref{ParamCond1}$ -  $\ref{ParamCond3}$ can be extended to other Machine-Learning methods than the Random Forest method. But before shortly introducing some considered boosting methods in this paper, the gap of using the Random Forest method in quantile regression is closed. A theoretical work of this can be found in \cite{meinshausen2006quantile} and is available in \textsf{R} through the package \texttt{quantregForest}.  
\begin{enumerate}
    \item \textbf{Quantile Regression Forest}.  The key idea is to   approximate the conditional distribution function $F(y| \boldsymbol{X} = \boldsymbol{x}) = \mathbb{P}[Y \le y | \boldsymbol{X} = \boldsymbol{x}]$, which itself can be rewritten into $\mathbb{E}[ \mathds{1} \{Y \le y \} |  \boldsymbol{X} = \boldsymbol{x} ]$ as an analgon to $m(\boldsymbol{x}) = \mathbb{E}[Y | \boldsymbol{X} = \boldsymbol{x}]$. The target variable then changes to $\Tilde{Y}_ i = \mathds{1}\{ Y_i \le y \} $ and a Random Forest method is constructed on the initial $\mathcal{D}_n$. The obtained weights from the trained Random Forest are used to approximate $F(y | \boldsymbol{X} = \boldsymbol{x})$ with the modified response variables  $\{ \Tilde{Y}_i \}_{i}$. The Random Forest approximation $\widehat{F}_n(y | \boldsymbol{X} = \boldsymbol{x})$ of $F(y | \boldsymbol{X} = \boldsymbol{x})$ is then used to compute quantiles for $\{ Y(\boldsymbol{x}_0) \} $ by considering
    \begin{align}
        \widehat{Q}_n( \boldsymbol{x}_0, \alpha) &= \inf\{y : \widehat{F}_n(y | \boldsymbol{X}_0 = \boldsymbol{x}_0) \ge \alpha  \}.
    \end{align}
    This would result into a type-III prediction interval of the form $[\widehat{Q}_n(\boldsymbol{x}_0, \alpha/2), \widehat{Q}_n(\boldsymbol{x}_0, 1 - \alpha/2)]$, whose validity is proven to hold asymptotically under rather strong assumptions such as the Lipschitz-continuity of the regression function $m$. Practical drawbacks are given by additional assumptions on the Random Forest method used in obtaining the quantiles $\widehat{Q}_n( \boldsymbol{x}_0, \alpha)$, which are practically not given when using Breiman's Random Forest method. These include assumptions on the tree construction process such as the regularization of the node-size in each tree. 
    \item \textbf{Stochastic Gradient Boosting Method.} The aim of boosting is to combine weak learners such as decision trees in an additive fashion in order to obtain a strong learner, see e.g. \cite{friedman2002stochastic}. In its core, it conducts the gradient descent method on the tree parameter space  to find the best parameter settings for fitting the next tree in an additive aggregation style. Within the regression context, the used squared error loss as the optimizing functions yields to residuals from the previous iteration step being fitted to the considered covariates. The randomness applies at every iteration step, where the gradient is not computed on the whole set $\mathcal{D}_n$, but on a random subset. Consistency results such as in $\ref{ParamCond1}$ and $\ref{ParamCond3}$ on this type of algorithm are not directly given, but for the general boosting method without random sampling in each iteration (see e.g. \cite{buehlmann2006boosting}). Therein, regression functions are assumed to be linear making theoretical transfer for obtaining prediction intervals more difficult. 
    \item \textbf{XGBoost}. The algorithm belongs to the general class of boosting methods developed in \cite{chen2016xgboost}, but showed competitive and favourable prediction results for a variety of data examples. Several additional features such as scalable end-to-end tree boosting system with a modified tree penalization, a proportional shrinking of leaf nodes as well as additional randomization parameters have been implemented for the general purpose to increase prediction accuracy compared to the traditional gradient boosting machine, while maintaining low space and time complexity. In this work, the \textsf{R}-package \texttt{xgboost} is used.  
\end{enumerate}

The verification of the conditions $\ref{ParamCond1}$ - $\ref{ParamCond3}$ for the stochastic gradient boosting method and the XGBoost method is not part of this work due to the extensive theoretical work involved in establishing consistency results for these type of learners. Instead, the parametric and non-parametric prediction intervals of the form given in $(\ref{Param_Interval})$ and $(\ref{NonParam_Interval})$ are used by substituting the corresponding learning method in $\widehat{m}_n$ and therefore also in $\widehat{D}_{n, \alpha}$ with the stochastic gradient tree booster resp. the XGBoost. 

\section{Simulation Design} \label{Sec:Simulation}

It is interesting to know how the different prediction intervals behave in practical problems. For this reason, an extensive simulation study is conducted in a Monte-Carlo based fashion focusing on three aspects regarding prediction uncertainty:
\begin{enumerate}[i)]
    \item Correct (asymptotic) coverage rates for type-I, type-II, type-III and type-IV prediction intervals, while an emphasis should be placed on the type-IV interval for practical usage. 
    \item Obtaining narrow intervals through the computation of interval lengths for the corresponding type of prediction interval. 
    \item Exhausting interval performance in terms of coverage rates and interval lengths by deviating from the theoretical assumptions given in $\ref{Ass1}$ -  $\ref{Ass4}$. Especially considering residual distributions that are non-normal with potential heavy tails and correlated features. 
\end{enumerate}

The \textit{signal-to-noise ratio} can be considered as a potential source of distortion regarding interval quality. This, because prediction interval coverage and length can be affected by the signal strength coming from the regression function and the additional noise arising from the residuals through its variance $\sigma^2$. This source of distortion in Random Forest models have been identified in \cite{ramosaj2019asymptotic}, for example. It is theoretically given by 
\begin{align}\label{SigNoiseRatio}
SN &= \frac{Var(m(\boldsymbol{X}))}{\sigma^2}, 
\end{align}
and is controlled in this study through the consideration of $SN \in \{0.5, 1, 3 \}$. One refers to noisy data, if $SN < 1$, while small noise data is present, if $SN>1$. Various regression functions have been considered for a $p = 10$ dimensional covariate space. Letting $\boldsymbol{\beta}_0 = [2,4,2,-3,1,7,-4,0,0,0]^\top$ be the regression coefficient, the following regression functions are then taken into account for $i = 1, \dots, n$: 
\begin{enumerate}
\item The \textbf{linear case} described by the relation $m(\boldsymbol{x}_i) = \boldsymbol{x}_i^\top \boldsymbol{\beta}_0$. \label{LinearModel}
\item The \textbf{polynomial case} given by $m(\boldsymbol{x}_i) = \sum\limits_{j = 1}^p \beta_{0, j} x_{i, j}^j$. \label{PolyModel}
\item The \textbf{trigonometric case} using $m(\boldsymbol{x}_i) = 2 \cdot sin( \boldsymbol{x}_i^\top \boldsymbol{\beta}_0 + 2 )$. \label{TrigonoModel}
\item The \textbf{non-continuous case} through the consideration of 
	\begin{align*}
			m(\mathbf{x}_i) &= 	
		\begin{cases}
			\beta_{0,1}x_{i,1} + \beta_{0,2}x_{i, 2} + \beta_{0,3} x_{i, 3}, &\text{ if } x_{i,3} > 0.5 \\
			\beta_{0,4}x_{i, 4} + \beta_{0,5}x_{i, 5} + 3 &\text{ if } x_{i,3} \le 0.5.
		\end{cases}
	\end{align*}\label{NonContModel}
\end{enumerate}
Note that the features $\{ \boldsymbol{X}_{i} \}_{i = 1}^n$ are iid generated using a multivariate normal distribution  and the inverse rule to transform the range back to the hyper-rectangular unit sphere $[0,1]^p$. This way, it is allowed to have potential dependencies among the features. Hence, one first generate $\widetilde{\boldsymbol{X}}_1 = [\widetilde{X}_{11}, \dots, X_{p1}]^\top \sim N_p(\boldsymbol{0}, \boldsymbol{\Sigma})$ and transform each feature back to the unit sphere through $\boldsymbol{X}_1 = [F_1(\widetilde{X}_{11}), \dots, F_p(\widetilde{X}_{p1})]^\top \in [0,1]^p$. $F_j$ denotes the normal distribution function with parameters $[\mu_X, \sigma_X^{2} ]^\top = [0, \boldsymbol{\Sigma}_{jj}]^\top$. Various covariance structures are considered allowing the analysis of potential dependence effects on covariates. This includes positive auto-regressive, negative-autoregressive, compound symmetric, linear decreasing Toeplitz and identity structures, where $\boldsymbol{I}_p$ denotes the identity matrix and $\boldsymbol{J}_p = \boldsymbol{1}_p \boldsymbol{1}_p^\top$ the matrix of ones: 
	\begin{align}\label{CovMatrices}
	\begin{aligned}
	\boldsymbol{\Sigma}_1 &= \{(-0.5)^{|i-j|} \}_{1 \le i,j \le p},   \\       \boldsymbol{\Sigma}_3 &= \boldsymbol{I}_p + \boldsymbol{J}_p, 
	\end{aligned}
	&&
	\begin{aligned}
	\boldsymbol{\Sigma}_2  &= \{ 0.5^{|i-j|} \}_{1 \le i,j \le p},\\       \boldsymbol{\Sigma}_4 &=  \{ 1- 1/p|i - j| \}_{ 1 \le i,j \le p},
	\end{aligned}
	&&
	\begin{aligned}
	\boldsymbol{\Sigma}_5 = \boldsymbol{I}_p.
	\end{aligned}
	\end{align}

Regarding the residual distribution, four different distributions are taken into consideration such as the normal, the $t$-, the exponential-, the log-normal distribution. Their parameters are determined by fixing the signal-to-noise ratio in $(\ref{SigNoiseRatio})$ to its respective range $\{ 0.5, 1,5  \}$. For details on the parameter derivations based on the $SN$, see the supplementary material. Uisng $MC = 1,000$ Monte-Carlo iterations, depending on the type of interval as given in Definition \ref{PI_Definition}, the coverage rate $p_{\alpha}$  for an interval $\mathcal{I}_{n, \alpha}^{(i)}$ in the $i$-th simulation run is then estimated by 
\begin{align}
    \widehat{p}_{n;\alpha} &= \frac{1}{MC} \sum\limits_{i = 1}^{MC} \mathds{1}\{ Y(\boldsymbol{X}_0^{(i)}) \in \mathcal{I}_{n, \alpha}^{(i)} \}.
\end{align}
Fixing the nominal coverage rate at $\alpha = 0.05$ and considering sample sizes $n \in \{100, 500, 1000 \}$, simulation based results are presented in the next section. Note that the presented boxplots are taken over the used covariance structures and the considered residual distributions. For the weighted, finite-$M$ corrected residual variance estimator used in the prediction interval $\mathcal{C}_{n, 1 -  \alpha, M}^W$, an equal weighting scheme is considered leading to $\lambda_1 = \lambda_2 = 0.5$. In addition to the used Machine Learning methods, the prediction interval obtained from a linear model is taken into consideration in order to obtain a trivial benchmark. The interval is given by 
\begin{align}\label{TRUELinModel}
    \left[ \boldsymbol{x}_0^\top \widehat{\boldsymbol{\beta}}_n  - t_{n - rk(\boldsymbol{X}), 1- \alpha/2} \cdot  \widehat{\sigma}_n \left( 1 + \boldsymbol{x}_0^\top ( \boldsymbol{X}^\top \boldsymbol{X} )^{-1} \boldsymbol{x}_0 \right), \quad \boldsymbol{x}_0^\top \widehat{\boldsymbol{\beta}}_n  + t_{n - rk(\boldsymbol{X}), 1-\alpha/2} \cdot  \widehat{\sigma}_n \left( 1 + \boldsymbol{x}_0^\top ( \boldsymbol{X}^\top \boldsymbol{X} )^{-1} \boldsymbol{x}_0 \right)  \right],
\end{align}
where $\widehat{\sigma}_n =  \frac{1}{n - rk(\boldsymbol{X})} \boldsymbol{Y}[ \boldsymbol{I}_n - \boldsymbol{X}(\boldsymbol{X}^\top \boldsymbol{X})^{-1} \boldsymbol{X}^\top ] \boldsymbol{Y}$, $\boldsymbol{Y}= [Y_1, \dots, Y_n]^\top$ and $\boldsymbol{X}$ is the design matrix of the covariates $\{ \boldsymbol{X}_i \}_{i = 1}^n$ restructed in an $n \times (p+1)$ matrix allowing the inclusion of a model intercept. Under the assumption of a linear model and $\ref{Ass3}$ together with $\det(\boldsymbol{X}^\top \boldsymbol{X}) \neq 0$, the linear prediction interval should be a valid type-IV interval. 

\section{Results} \label{Sec:SimResults}

The simulation results for a signal-to-noise ratio of $SN=1$ are presented. The other choices of $SN$ can be found in the Supplement on the Simulation Section. Note that the simulation results of the Stochastic Gradient Boosting Method and the XGBoost method are not presented. This, because of their relatively poor performance regarding coverage rates. On average, they resulted into rates being approximately $0.03$ while interval length was comparably large.  However, the results based on the derived parametric Random Forest prediction intervals are completely different.\\

\begin{figure*}[h!]
	\centering
	\includegraphics[width=5in]{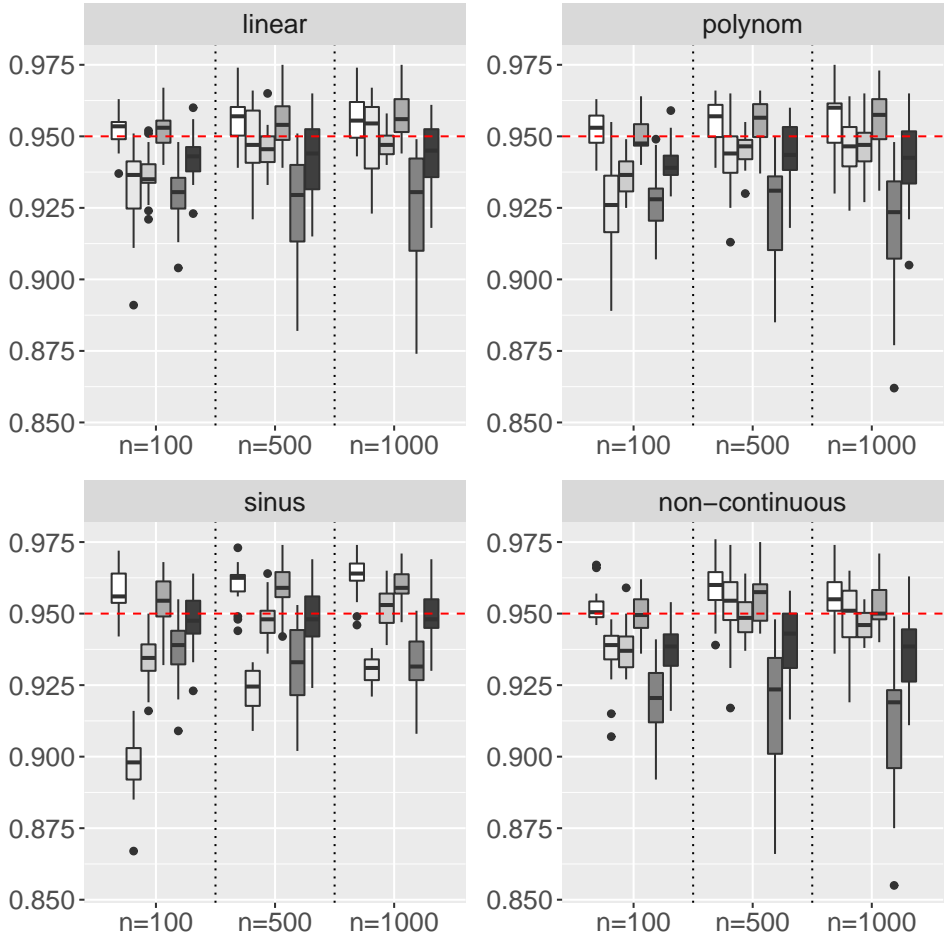}
	\caption{Simulation results on the \textbf{coverage rate} for the \textbf{type-I} prediction interval using a signal-to-noise ratio of $SN = 1$ with $MC = 1,000$ Monte-Carlo iterations. The arranged boxplots in every sample size segment corresponds to the following methods: (from left to right)
	\textit{Linear Model}, \textit{Quantile Regression Forest},
	\textit{Random-Forest with Empirical Quantiles},
	\textit{Random Forest with simple residual variance estimator as in $\mathcal{C}_{n, 1-\alpha, M}$},
	\textit{Random-Forest with finite-$M$-corrected estimator as in $\mathcal{C}_{n, 1- \alpha, M}^{Cor}$ }, \textit{Random Forest with weighted residual variance as in $\mathcal{C}_{n, 1- \alpha, M}^{W}$. } }
	\label{T1_Coverage_SN1}
\end{figure*}

Starting with the coverage rate of the type-I prediction interval given in Figure \ref{T1_Coverage_SN1} with $SN=1$, the non-parametric Random Forest prediction interval based on empirical quantiles (NP-RF-EQ) and the parametric Random Forest prediction interval with the usual sampling variance (P-RF) yielded even better coverage rates under the linear model than the interval in $(\ref{TRUELinModel})$. The parametric Random Forest prediction interval with finite-$M$-correction (P-RF-MCor) was more conservative under the type-I prediction interval scheme leading to less accurate, but shorter prediction intervals. The weighted, finite-$M$ corrected prediction interval (P-RF-W) smoothed out the effect by being more accurate than P-RF-MCor resulting to almost perfect coverage rates on average for the trigonometric regression function. As preliminary suggested, the performance of P-RF method in terms of accurate coverage was slightly  more liberal, but still very competitive. A potential reason for this is the slight overestimation of the residual variance due to the finite choice of $M$. The P-RF-W method balanced out this effect. In type-IV prediction intervals, one will see that this effect turns out to be even stronger. Comparing the interval coverage among the Random-Forest methods, NP-RF-EQ increased in coverage quality, especially when the sample size increased across the different regression function. For small to moderate sample sizes, the P-RF interval was on average more accurate. In summary, both, the NP-RF-EQ interval and the P-RF interval yielded under a neutral noise setting ($SN = 1$) competitive coverage results. The Quantile Regression Forest (QRF), however, had difficulties in obtaining correct coverage rates, especially for small samples and non-linear regression functions, being sometimes even worse than the interval of the linear model. Interesting is also the effect of non-normal residuals on the coverage rate. Both, the NP-RF-EQ and the P-RF method yielded more robust results under the type-I scheme. The P-RF-MCor and the QRF method, however, were more volatile. The reason for the P-RF-MCor method can be the obtained error bound itself, which is based on the normal assumption. \\

\begin{figure*}[h!]
	\centering
	\includegraphics[width=5in]{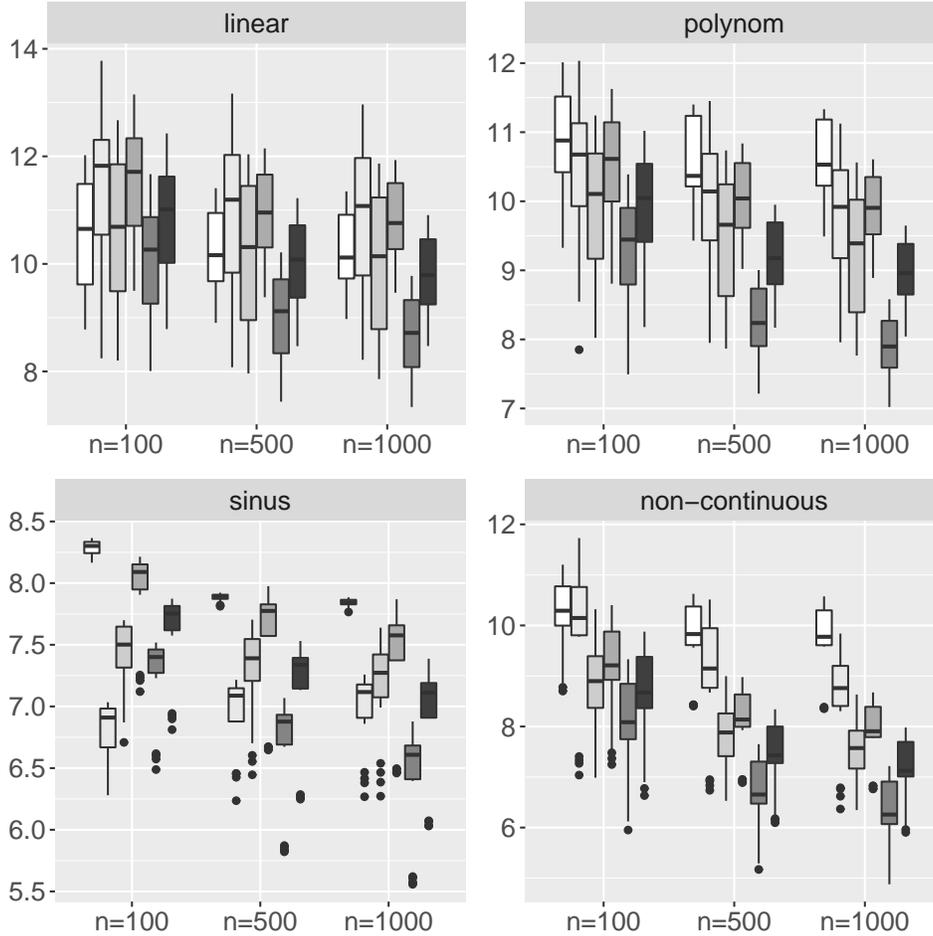}
	\caption{Simulation results on the \textbf{interval length} for the \textbf{type-I} prediction interval using a signal-to-noise ratio of $SN = 1$ with $MC = 1,000$ Monte-Carlo iterations. The arranged boxplots in every sample size segment corresponds to the following methods: (from left to right)
	\textit{Linear Model}, \textit{Quantile Regression Forest},
	\textit{Random-Forest with Empirical Quantiles},
	\textit{Random Forest with simple residual variance estimator as in $\mathcal{C}_{n, 1-\alpha, M}$},
	\textit{Random-Forest with finite-$M$-corrected estimator as in $\mathcal{C}_{n, 1- \alpha, M}^{Cor}$ }, \textit{Random Forest with weighted residual variance as in $\mathcal{C}_{n, 1- \alpha, M}^{W}$. } }
	\label{T1_Length_SN1}
\end{figure*}

Regarding type-I prediction interval length, the P-RF-MCor and the P-RF-W  resulted into the most narrow intervals, while the NP-RF-EQ interval and the P-RF were slightly larger being on average similar in length.  This effect can be extracted from Figure \ref{T1_Length_SN1}. Due to the more stringent assumption on the residual distribution given in $\ref{Ass3}$ for the parametric intervals, the interval length got reduced accordingly compared to the NP-RF-EQ prediction interval, making all three parametric intervals more competitive, especially the P-RF method. The prediction interval of a linear model and the QRF method indicated much wider interval lengths. Taking both, the coverage rate and the interval length under the $SN=1$ scenario together, the P-RF method was competitive leading to accurate coverage rates while having narrow interval lengths. \\

\begin{figure*}[h!]
	\centering
	\includegraphics[width=5in]{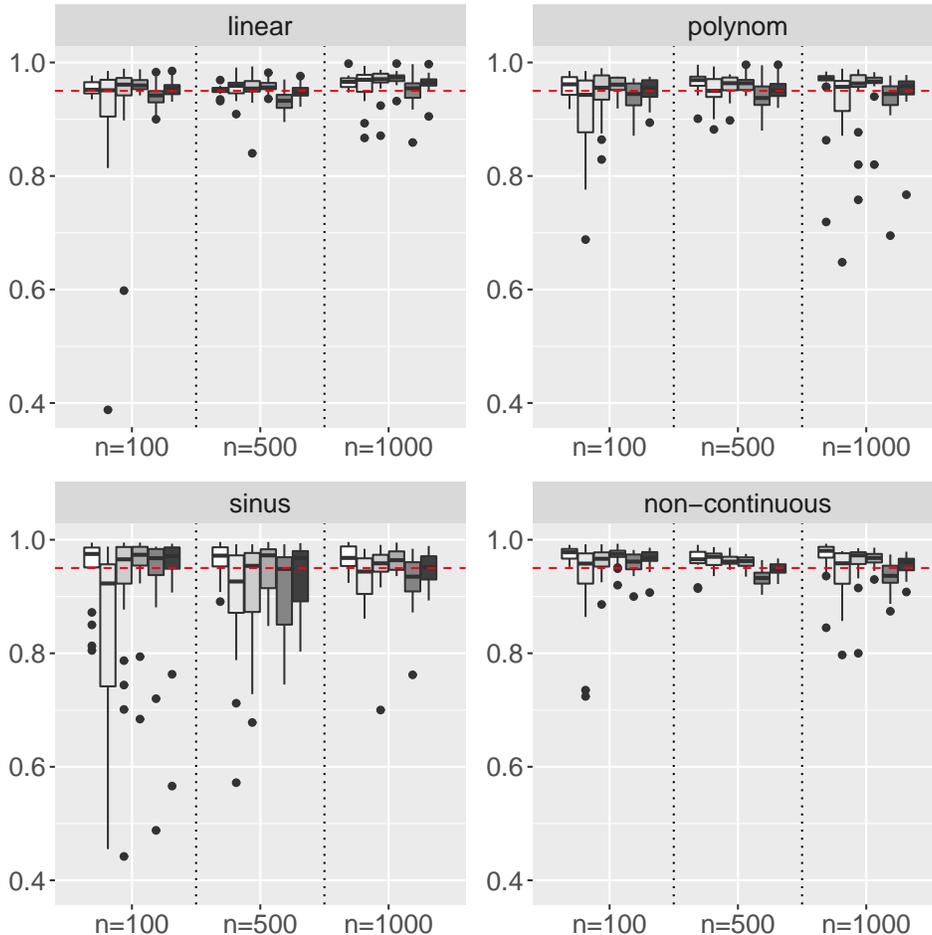}
	\caption{Simulation results on the \textbf{coverage rate} for the \textbf{type-IV} prediction interval using a signal-to-noise ratio of $SN = 1$ with $MC = 1,000$ Monte-Carlo iterations. The arranged boxplots in every sample size segment corresponds to the following methods: (from left to right)
	\textit{Linear Model}, \textit{Quantile Regression Forest},
	\textit{Random-Forest with Empirical Quantiles},
	\textit{Random Forest with simple residual variance estimator as in $\mathcal{C}_{n, 1-\alpha, M}$},
	\textit{Random-Forest with finite-$M$-corrected estimator as in $\mathcal{C}_{n, 1- \alpha, M}^{Cor}$}, \textit{Random Forest with weighted residual variance as in $\mathcal{C}_{n, 1- \alpha, M}^{W}$.} }
	\label{T4_Coverage_SN1}
\end{figure*}

Returning to the stronger type-IV prediction interval given in Figure \ref{T4_Coverage_SN1}, the proposed P-RF-MCor prediction interval yielded strong coverage rates lying slightly below $0.95$ on average. When combining it with the P-RF interval yielding to the weighted, parametric interval P-RF-W, almost accurate coverage results could be obtained throughout the different sample sizes and the considered regression methods. Therefore, the P-RF, the P-RF-MCor and the P-RF-W method yielded the most competitive coverage results for type-IV prediction intervals, while the P-RF-W was even more accurate under this scheme. The QRF interval turned out to be more accurate under the type-IV prediction interval than under the type-I interval, but indicated a higher coverage volatility being thus more sensitive to non-normal residuals and correlated features than their paremetric counterparts P-RF, P-RF-MCor and P-RF-W. The non-parametric Random Forest interval NP-RF-EQ performed comparable to the P-RF interval, but was more liberal than the  latter. The result indicate that the correction for a finite choice of $M$ is not neglectable and can significantly improve type-IV prediction interval coverage.

\begin{figure*}[h!]
	\centering
	\includegraphics[width=5in]{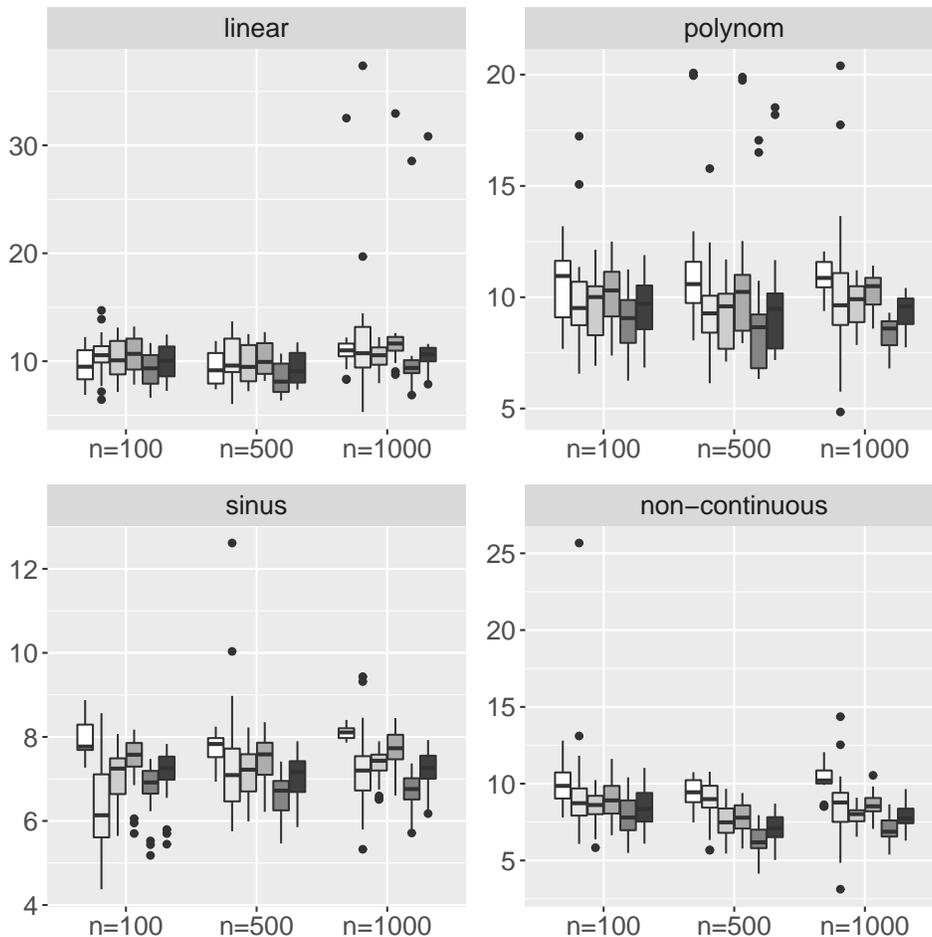}
	\caption{Simulation results on the \textbf{interval length} for the \textbf{type-IV} prediction interval using a signal-to-noise ratio of $SN = 1$ with $MC = 1,000$ Monte-Carlo iterations. The arranged boxplots in every sample size segment corresponds to the following methods: (from left to right)
	\textit{Linear Model}, \textit{Quantile Regression Forest},
	\textit{Random-Forest with Empirical Quantiles},
	\textit{Random Forest with simple residual variance estimator as in $\mathcal{C}_{n, 1-\alpha, M}$},
	\textit{Random-Forest with finite-$M$-corrected estimator as in $\mathcal{C}_{n, 1- \alpha, M}^{Cor}$}, \textit{Random Forest with weighted residual variance as in $\mathcal{C}_{n, 1- \alpha, M}^{W}$.} }
	\label{T4_Length_SN1}
\end{figure*}

The type-IV interval length results are summarized in Figure \ref{T4_Length_SN1} for a signal-to-noise ratio of $SN=1$. Also here, the P-RF-MCor and the P-RF-W interval resulted into the most narrow interval in almost all considered scenario. The P-RF method also indicated narrow interval lengths, but the correction due to a finite choice of $M$ indicated a positive trend in prediction interval quality. Regarding prediction interval length the P-RF method performed comparable to the NP-RF-EQ method. Combining the simulated coverage rates and the type-IV prediction interval length, the P-RF-W interval belonged to the most competitive one yielding almost accurate coverage rates and comparably narrow interval lengths. Its unweighted counterpart, the P-RF-MCor method, was in total competitive, too, being even more narrow.

The results regarding type-II and type-III prediction intervals for $SN=1$ are moved to the supplement. For both types of intervals, especially the type-III prediction interval, the P-RF-W method yielded on average very competitive coverage results, while also being comparably narrow. Under the type-II scheme, the method was also competitive, but for an increased sample size. The NP-RF-EQ interval, however, performed worse than their parametric counterparts, especially when the sample size was comparably small. The effect reduced with an increased sample size, but was more sensitive towards the chosen regression function. Similarly to the previous results, the QRF interval was more volatile, especially under the type-III scheme. Regarding interval length similar results as under the type-IV and type-I scheme could be observed. Especially the P-RF-MCor and the P-RF-W intervals yielded the tightest intervals. Combigning these finidings with the type-III coverage rates, the P-RF-W method resulted into competitive interval quality.

The simulation study also revealed that the signal-to-noise ratio is not neglectable. For small values of $SN$, such as $SN = 0.5$, which indicates the presence of noisy data due to the relation $Var(m(\boldsymbol{X})) < \sigma^2$,  the prediction interval quality in terms of coverage and length turned even more accurate for the P-RF-W interval for all prediction interval types. Especially the more conservative type-IV version indicated preferable results for the P-RF-W and the P-RF-MCor method. The results can be found in the Simulation Section of the Supplement. Regarding the performance of the NP-RF-EQ interval, the quality in terms of length and accurate coverage suffered from noisy data. An increased sample size for this method turned prediction interval performance into a positive direction for all type of intervals according to Definition \ref{PI_Definition}. For an increased signal-to-noise ratio such as $SN=3$, the interval quality of P-RF-MCor and P-RF-W suffered under a more conservative estimation of the residual variance leading to a more conservative estimation of the coverage rate for the type-I, type-II and type-III prediction interval. This can also be seen from the simulation results on the prediction interval length. Therein, the P-RF-MCor method yielded the shortest intervals on average penalizing the estimation of the residual variance stronger than their counterpart. The reason lies in the relation that $Var(m(\boldsymbol{X}))>\sigma^2$ making the effect of a finite-$M$ bias on the final output prediction stronger and impacting therefore prediction interval quality stronger. However, for these types of intervals (type-I, type-II and type-III), the parametric and non-penalizing method P-RF and NP-RF-EQ under $SN>1$ yielded the most favourable results. However, regarding the type-IV prediction interval, the P-RF-W method remained among the best method under the aspect of correct coverage rates and comparably narrow interval length. In this case, the NP-RF-EQ method and the P-RF method performed comparably well, resulting into interval lengths being similar.

\section{Conclusions} \label{Sec:Conclusion}

The construction of trustworthy and interpretable  models is a general issue in statistical machine learning. Beside accurate point predictions the quantification of uncertainty is of specific importance. In particular, the construction of statistically valid prediction intervals for the quantification of uncertainty regarding future point prediction contributes to the problem of constructing trustworthy machines. Focusing on the Random Forest method and other machine learning algorithms, formal conditions have been set to guarantee correct (asymptotic) coverage rates under parametric interval types. Therein, different types of prediction intervals have been considered mainly differing in the conditioning of the underlying probability measure. While type-I and type-II prediction intervals also average over unseen prediction points, their practical interpretability and usability are more restrictive. Therefore, type-III and type-IV prediction intervals, which keep unseen observations fixed, are easier to interpret and of practical importance. Based on this distinction, the problem of obtaining correct prediction intervals reduced to the issue of consistently estimating residual variance in regression learning methods. Therein, several estimators have been proposed and central sources of distortions have been identified such as the bias due to the finite choice of decision trees in the forest (finite-$M$-bias) and the signal-to-noise ratio $(SN)$. The proposed parametric prediction intervals delivered accurate coverage rates and narrow interval lengths. Especially the interval based on a weighted residual variance scheme resulted into favorable coverage rates in obtaining correct point-wise prediction intervals, such as the type-III and type-IV interval.

For noisy $(SN>0.5)$ and neutral noise containing data $(SN=1)$, the derived parametric intervals such as the weighted finite-$M$-corrected version or the interval with a simple Random Forest based residual variance estimator, delivered competitive results. They outperformed the Random Forest prediction interval based on empirical residuals (see \cite{zhang2019random}) and the Quantile Regression Forest (see \cite{meinshausen2006quantile}). For larger signal-to-noise ratios $(SN=3)$, the finite-$M$ corrected version penalized residual variance estimators stronger leading to more conservative coverage rates for type-I, type-II and type-III prediction intervals. In this case, the uncorrected Random Forest based parametric prediction interval was more accurate. However, under the type-IV prediction interval, the weighted finite-$M$ corrected version delivered in almost all scenarios accurate coverage rates while maintaining narrow interval length even under non-normal residual distributions and dependent covariate structures. Therefore, the strong type-IV prediction interval quality substantially increased when using the weighted, finite-$M$ corrected version with equal weights. The results also leave optimistic space to even enhanced results for the type-I and type-II prediction interval, if data dependent weighting as a future research field is implemented for the weighted, finite-$M$ corrected Random Forest prediction interval. \\

Other machine learning methods such as prediction intervals based on the stochastic gradient tree boosting method or the XGBoost method were not able to deliver accurate coverage rates in the simulation study. Future research work will cover the involvement of missing values in the covariates and its impact in accurate prediction interval coverage and length.

\section*{Acknowledgement}

The author's work is funded by the Ministry of of Culture and Science of the state of NRW (MKW NRW) through the research grand programme \textit{KI-Starter}. The authors gratefully acknowledge the computing time provided on the Linux HPC cluster at Technical University Dortmund (LiDO3), partially funded in the course of the Large-Scale Equipment Initiative by the German Research Foundation (DFG) as project 271512359. Moreover, the author likes to thank Markus Pauly for valuable thoughts and vital discussions on Random Forest related issues in uncertainty quantification. 
 
\newpage

\bibliography{BibFile}

\begin{thebibliography}{10}
\expandafter\ifx\csname url\endcsname\relax
  \def\url#1{\texttt{#1}}\fi
\expandafter\ifx\csname urlprefix\endcsname\relax\def\urlprefix{URL }\fi
\expandafter\ifx\csname href\endcsname\relax
  \def\href#1#2{#2} \def\path#1{#1}\fi

\bibitem{buhlmann2003bagging}
P.~L. B{\"u}hlmann, Bagging, subagging and bragging for improving some
  prediction algorithms, in: Research Report/Seminar f{\"u}r Statistik,
  Eidgen{\"o}ssische Technische Hochschule (ETH), Vol. 113, Seminar f{\"u}r
  Statistik, Eidgen{\"o}ssische Technische Hochschule (ETH), Z{\"u}rich, 2003.

\bibitem{prasad2006newer}
A.~M. Prasad, L.~R. Iverson, A.~Liaw, {Newer Classification and Regression Tree
  Techniques: Bagging and Random Forests for Ecological Prediction}, Ecosystems
  9~(2) (2006) 181--199.

\bibitem{jiang2007mipred}
P.~Jiang, H.~Wu, W.~Wang, W.~Ma, X.~Sun, Z.~Lu, {MiPred: classification of real
  and pseudo microRNA precursors using random forest prediction model with
  combined features}, Nucleic Acids Research 35~(2) (2007) W339--W344.

\bibitem{pham2018bagging}
B.~T. Pham, D.~T. Bui, I.~Prakash, {Bagging based Support Vector Machines for
  spatial prediction of landslides}, Environmental Earth Sciences 77~(4) (2018)
  1--17.

\bibitem{wang2018random}
Z.~Wang, Y.~Wang, R.~Zeng, R.~S. Srinivasan, S.~Ahrentzen, {Random Forest based
  hourly building energy prediction}, Energy and Buildings 171 (2018) 11--25.

\bibitem{breiman2001random}
L.~Breiman, Random {F}orests, Machine Learning 45~(1) (2001) 5--32.

\bibitem{diaz2006gene}
R.~D{\'\i}az-Uriarte, S.~A. De~Andres, Gene {S}election and {C}lassification of
  {M}icroarray {D}ata using {R}andom {F}orest, BMC Bioinformatics 7~(3) (2006).
\newblock \href {https://doi.org/https://doi.org/10.1186/1471-2105-7-3}
  {\path{doi:https://doi.org/10.1186/1471-2105-7-3}}.

\bibitem{genuer2010variable}
R.~Genuer, J.-M. Poggi, C.~Tuleau-Malot, Variable {S}election using {R}andom
  {F}orests, Pattern Recognition Letters 31~(14) (2010) 2225--2236.

\bibitem{ramosaj2019asymptotic}
B.~Ramosaj, M.~Pauly, {Asymptotic Unbiasedness of the Permutation Importance
  Measure in Random Forest Models}, arXiv preprint arXiv:1912.03306 (2019).

\bibitem{wager2014asymptotic}
S.~Wager, S.~Athey, {Estimation and Inference of Heterogeneous Treatment
  Effects using Random Forests}, Journal of the American Statistical
  Association 113~(523) (2018) 1228--1242.

\bibitem{scornet2015consistency}
E.~Scornet, G.~Biau, J.-P. Vert, Consistency of {R}andom {F}orests, The Annals
  of Statistics 43~(4) (2015) 1716--1741.

\bibitem{mentch2016quantifying}
L.~Mentch, G.~Hooker, Quantifying {U}ncertainty in {R}andom {F}orests via
  {C}onfidence {I}ntervals and {H}ypothesis {T}ests, The Journal of Machine
  Learning Research 17~(1) (2016) 841--881.

\bibitem{zhou2019v}
Z.~Zhou, L.~Mentch, G.~Hooker, {$V$-statistics and Variance Estimation}, arXiv
  preprint arXiv:1912.01089 (2019).

\bibitem{wager2014confidence}
S.~Wager, T.~Hastie, B.~Efron, Confidence {I}ntervals for {R}andom {F}orests:
  The {J}ackknife and the {I}nfinitesimal {J}ackknife, Journal of Machine
  Learning Research 15~(1) (2014) 1625--1651.

\bibitem{adadi2018peeking}
A.~Adadi, M.~Berrada, {Peeking Inside the Black-Box: A Survey on Explainable
  Artificial Intelligence (XAI)}, IEEE Access 6 (2018) 52138--52160.

\bibitem{dunson2018statistics}
D.~B. Dunson, {Statistics in the big data era: Failures of the machine},
  Statistics \& Probability Letters 136 (2018) 4--9.

\bibitem{guidotti2018survey}
R.~Guidotti, A.~Monreale, S.~Ruggieri, F.~Turini, F.~Giannotti, D.~Pedreschi,
  {A Survey of Methods for Explaining Black Box Models}, ACM Computing Surveys
  (CSUR) 51~(5) (2018) 1--42.

\bibitem{molnar2020InterpretableMachine}
M.~Christoph, Interpretable Machine Learning, Lulu.com, 2020.

\bibitem{coulston2016approximating}
J.~W. Coulston, C.~E. Blinn, V.~A. Thomas, R.~H. Wynne, {Approximating
  Prediction Uncertainty for Random Forest Regression Models}, Photogrammetric
  Engineering \& Remote Sensing 82~(3) (2016) 189--197.

\bibitem{calvino2020random}
A.~Calvi{\~n}o, {On Random-Forest-Based Prediction Intervals}, in:
  International Conference on Intelligent Data Engineering and Automated
  Learning, Springer, 2020, pp. 172--184.

\bibitem{roy2020prediction}
M.-H. Roy, D.~Larocque, {Prediction Intervals with Random Forests}, Statistical
  Methods in Medical Research 29~(1) (2020) 205--229.

\bibitem{meinshausen2006quantile}
N.~Meinshausen, G.~Ridgeway, {Quantile Regression Forests.}, Journal of Machine
  Learning Research 7~(6) (2006).

\bibitem{zhang2019random}
H.~Zhang, J.~Zimmerman, D.~Nettleton, D.~J. Nordman, {Random Forest Prediction
  Intervals}, The American Statistician (2019).

\bibitem{ramosaj2019consistent}
B.~Ramosaj, M.~Pauly, {Consistent estimation of residual variance with random
  forest Out-Of-Bag errors}, Statistics \& Probability Letters 151 (2019)
  49--57.

\bibitem{ramosaj2020Dissertation}
B.~Ramosaj, {Analyzing Consistency and Statistical Inference in Random Forest
  Models.}, Dissertation. Repositorium der TU Dortmund (2020).

\bibitem{lei2018distribution}
J.~Lei, M.~G’Sell, A.~Rinaldo, R.~J. Tibshirani, L.~Wasserman,
  {Distribution-Free Predictive Inference for Regression}, Journal of the
  American Statistical Association 113~(523) (2018) 1094--1111.

\bibitem{friedman2002stochastic}
J.~H. Friedman, {Stochastic Gradient Boosting}, Computational Statistics \&
  Data Analysis 38~(4) (2002) 367--378.

\bibitem{buehlmann2006boosting}
P.~Buehlmann, {Boosting for High-Dimensional Linear Models}, The Annals of
  Statistics 34~(2) (2006) 559--583.

\bibitem{chen2016xgboost}
T.~Chen, C.~Guestrin, {XGBoost: A Scalable Tree Boosting System}, in: KDD '16:
  Proceedings of the 22nd ACM SIGKDD International Conference on Knowledge
  Discovery and Data Mining, 2016, pp. 785--794.

\end{thebibliography}


\begin{thebibliography}{1}
\expandafter\ifx\csname url\endcsname\relax
  \def\url#1{\texttt{#1}}\fi
\expandafter\ifx\csname urlprefix\endcsname\relax\def\urlprefix{URL }\fi
\expandafter\ifx\csname href\endcsname\relax
  \def\href#1#2{#2} \def\path#1{#1}\fi

\bibitem{ramosaj2019consistent}
B.~Ramosaj, M.~Pauly, {Consistent estimation of residual variance with random
  forest Out-Of-Bag errors}, Statistics \& Probability Letters 151 (2019)
  49--57.

\bibitem{scornet2016asymptotics}
E.~Scornet, On the {A}symptotics of {R}andom {F}orests, Journal of Multivariate
  Analysis 146 (2016) 72--83.

\end{thebibliography}

\end{document}